\pdfoutput=1
\documentclass{article}
\usepackage[utf8]{inputenc}
\usepackage[letterpaper]{geometry}
\usepackage[parfill]{parskip}
\PassOptionsToPackage{numbers, compress}{natbib}
\usepackage{natbib}
\usepackage{xr-hyper,refcount}
\usepackage{graphicx}
\usepackage[T1]{fontenc}    
\usepackage[colorlinks, citecolor=blue, urlcolor=blue, linkcolor=blue]{hyperref}       
\usepackage{url}            
\usepackage{booktabs}       
\usepackage{amsfonts}       
\usepackage{nicefrac}       
\usepackage{microtype}      
\usepackage[parfill]{parskip}
\usepackage{algorithm,algorithmic}
\usepackage{amsmath,amsthm,amssymb,bbm}
\usepackage{mathtools}
\usepackage{cases}
\usepackage{comment}
\usepackage{subcaption}
\usepackage{algorithm,algorithmic}
\usepackage{color}
\usepackage{xcolor}
\usepackage{appendix}
\usepackage{xspace}
\usepackage[shortlabels]{enumitem}
\usepackage[english]{babel}
\usepackage{authblk}
\usepackage{libertine}
\usepackage{libertinust1math}
\usepackage{bm}
\usepackage{arydshln}
\usepackage{makecell}
\usepackage{multirow}
\usepackage{framed}
\usepackage[font=small]{caption}
\usepackage[scientific-notation=true]{siunitx}
\usepackage{wrapfig}
\usepackage{lipsum}
\usepackage{sidecap}
\usepackage{pifont}
\usepackage{tikz}
\theoremstyle{plain}
\usepackage{fixltx2e}
\usepackage[flushleft]{threeparttable}
\usepackage{diagbox}

\newcommand{\namel}{\textsc{Noisy Identity MLP}}
\newcommand{\names}{\textsc{NI-MLP}}

\DeclareMathOperator*{\argmin}{arg\,min}

\definecolor{gg}{RGB}{15,150,15}
\definecolor{rr}{RGB}{230,45,45}

\makeatletter
\def\maketag@@@#1{\hbox{\m@th\normalfont\normalsize#1}}
\makeatother

\usepackage{amsmath,amsfonts,bm}

\def\eqref#1{eq~(\ref{#1})}

\def\1{\bm{1}}

\DeclareMathAlphabet{\mathsfit}{\encodingdefault}{\sfdefault}{m}{sl}
\SetMathAlphabet{\mathsfit}{bold}{\encodingdefault}{\sfdefault}{bx}{n}

\usepackage{hyperref}

\usepackage{booktabs}       %
\usepackage{amsfonts}       %
\usepackage{nicefrac}       %
\usepackage{microtype}      %
\usepackage{subcaption}

\usepackage{enumitem}
\setlist{nolistsep}
\setlist[itemize]{noitemsep, topsep=0pt}

\usepackage{times}

\usepackage{graphicx} %
\usepackage{color}

\usepackage{array}
\usepackage{amssymb}
\usepackage{amsmath}
\usepackage{xspace}
\usepackage{fancyhdr}
\usepackage{comment}

\newcolumntype{H}{>{\setbox0=\hbox\bgroup}c<{\egroup}@{}}

\newcommand{\noaistats}[1]{}  %

\definecolor{darkgreen}{rgb}{0,0.4,0.0}
\definecolor{darkblue}{rgb}{0,0.1,0.3}
\definecolor{darkred}{rgb}{0.7,0.0,0.0}

\newcommand\blfootnote[1]{%
  \begingroup
  \renewcommand\thefootnote{}\footnote{#1}%
  \addtocounter{footnote}{-1}%
  \endgroup
}

\AtBeginDocument{%
  \addtolength\abovedisplayskip{-0.3\baselineskip}%
  \addtolength\belowdisplayskip{-0.3\baselineskip}%
}

\title{Multimodal Privacy-preserving Mood Prediction\\from Mobile Data: A Preliminary Study}

\author{%
  Terrance Liu$^{1\star}$, Paul Pu Liang$^{1\star}$, Michal Muszynski$^1$, Ryo Ishii$^1$, \vadjust{\vspace{2pt}}\nolinebreak\hspace{\fill}\linebreak David Brent$^2$, Randy P. Auerbach$^3$, Nicholas B. Allen$^4$, Louis-Philippe Morency$^1$\\
  $^1$School of Computer Science, Carnegie Mellon University\\
  $^2$Department of Psychiatry, University of Pittsburgh\\
  $^3$Department of Psychiatry, Columbia University\\
  $^4$Department of Psychology, University of Oregon\\
  \texttt{\{terrancl,pliang,morency\}@cs.cmu.edu}\\
}

\begin{document}

\maketitle

\begin{abstract}

Mental health conditions remain under-diagnosed even in countries with common access to advanced medical care. The ability to accurately and efficiently predict mood from easily collectible data has several important implications towards the early detection and intervention of mental health disorders. One promising data source to help monitor human behavior is from daily smartphone usage. However, care must be taken to summarize behaviors without identifying the user through personal (e.g., personally identifiable information) or protected attributes (e.g., race, gender). In this paper, we study behavioral markers or daily mood using a recent dataset of mobile behaviors from high-risk adolescent populations. Using computational models, we find that multimodal modeling of both text and app usage features is highly predictive of daily mood over each modality alone. Furthermore, we evaluate approaches that reliably obfuscate user identity while remaining predictive of daily mood. By combining multimodal representations with privacy-preserving learning, we are able to push forward the performance-privacy frontier as compared to unimodal approaches.\blfootnote{$^\star$first two authors contributed equally.}

\end{abstract}

\vspace{-1mm}
\section{Introduction}
\vspace{-1mm}

Mental illnesses can have a damaging permanent impact on communities, societies and economies all over the world. Individuals often do not realize they are at risk of mental disorders even when they have some mild symptoms. As a result, many are late in seeking professional help~\citep{thornicroft2016evidence}, particularly among adolescents where suicide is the second leading cause of death~\citep{curtin2019death}. In addition to deaths, 16$\%$ of high school students report having seriously suicidal thoughts each year, and 8$\%$ of them make one or more suicide attempts~\citep{cdc}.
Moreover, children with high irritability as well as depressive and anxious mood have a higher suicidal risk during adolescence compared with children with low symptoms~\citep{orri2018association}.
As a step towards adaptive interventions of suicidal ideations, intensive monitoring of behaviors via adolescents' use of smartphones may shed new light on the early risk of suicidal thoughts and ideations~\citep{nahum2018just}. While smartphones provide a valuable data source, one must take care to summarize behaviors without revealing user identities through personal (e.g., personally identifiable information) or protected attributes (e.g., race, gender) to potentially adversarial third-parties~\citep{sharma2018toward}. This form of anonymity is critical for implementing technologies in real-world scenarios.

Recent work in affective computing has begun to explore the potential in predicting mood and emotion from mobile data. In particular, they have found that typing patterns~\citep{cao2017deepmood,ghosh2017evaluating,huang2018dpmood,zulueta2018predicting}, self-reporting apps~\citep{suhara2017deepmood}, and wearable sensors~\citep{ghandeharioun2017objective,ghosh2017tapsense,sano2018identifying} are particularly predictive for affect. In addition, multimodal modeling of multiple sensors (e.g. wearable sensors and smartphone apps) was shown to further improve performance~\citep{jaques2017multimodal,taylor2017personalized}. While current work primarily relies on self-report apps for \textit{long-term} assessments of suicide risk~\citep{glenn2014improving}, our work investigates fine-grained text typed and apps used throughout the entire day as a signal to detect \textit{imminent} suicidal risk~\citep{franklin2017risk,large2017patient}, which is of critical clinical importance~\citep{glenn2014improving}.

Recent studies have also shown that private traits are predictable from digital records of human behavior~\citep{kosinski2013private}, which is dangerous especially when sensitive user data is concerned. As a result, in parallel to improving predictive performance, a recent focus has been on improving the privacy of real-world machine learning models through techniques such as differential privacy~\citep{dankar2012application,dankar2013practicing,dankar2012estimating} and federated learning~\citep{DBLP:journals/corr/McMahanMRA16,geyer2017differentially,liang2020think}, especially for healthcare data (e.g. EHRs~\citep{xu2019federated} and wearable devices~\citep{chen2020fedhealth}).

In this paper, as a step towards using \textit{multimodal privacy-preserving} mood prediction as fine-grained signals to detect \textit{imminent} suicidal risk, we analyze a recent dataset of mobile behaviors collected from adolescent populations at high suicidal risk.
With consent from participating groups, the dataset collects fine-grained features spanning online communication, keystroke patterns, and application usage. Participants are administered clinical interviews probing daily mood scores. Using this unique dataset, we first find that machine learning algorithms on extracted word-level and app history features provide a promising learning signal in predicting daily self-reported mood scores. Furthermore, we find that joint multimodal modeling of both text and app usage features allows models to \textit{contextualize} text typed in their corresponding apps which leads to better predictive performance over each modality alone. As a step towards privacy-preserving learning, we also evaluate approaches that obfuscate user identity while remaining predictive of daily mood. Finally, by combining multimodal contextualization with privacy-preserving learning, we are able to further push forward the performance-privacy frontier as compared to unimodal approaches.

\vspace{-1mm}
\section{Multimodal Mobile Dataset}
\vspace{-1mm}

Intensive monitoring of behaviors via adolescents' frequent use of smartphones may shed new light on the early risk of suicidal thoughts and ideations~\citep{nahum2018just}. Smartphones provide a valuable and natural data source with rich behavioral markers spanning online communication, keystroke patterns, and application usage. Learning these behavioral markers requires large datasets with diversity in participants, variety in features collected, and accuracy in annotations. As a step towards this goal, we have recently collected a dataset of mobile behaviors from high-risk adolescent populations with consent from participating groups.

We begin with a brief review of the data collection process. This data monitors adolescents spanning (a) recent suicide attempters (past 6 months) with current suicidal ideation, (b) suicide ideators with no past suicide attempts, and (c) psychiatric controls with no history of suicide ideation or attempts. Passive sensing data is collected from each participant’s smartphone across a duration of 6 months. Participants are administered clinical interviews probing for suicidal thoughts and behaviors (STBs) and self-report instruments regarding symptoms, and acute events (e.g., suicide attempts, psychiatric hospitalizations) are tracked weekly via a questionnaire. All users have given consent for their mobile device data to be collected and shared with us for research purposes.

\textbf{Labels:} Each day, users are asked to respond to the following question - ``In general, how have you been feeling over the last day?'' - with an integer score between 0 and 100, where 0 means very negative and 100 means very positive. To construct our prediction task, we discretized these scores into the following three bins - \textit{negative} (0-33), \textit{neutral} (34-66), \textit{positive} (67-100), which follows a class distribution of 12.43\%, 43.63\%, and 43.94\% respectively. For our 3-way classification task, participants with fewer than 50 daily self-reports were removed since such participants do not provide enough data to train an effective model. In total, our dataset consists of 1641 samples, consisting of data coming from 17 unique participants.

\textbf{Features:} For our study, we focused on keyboard data, which includes the time of data capture, the application used, and the text entered by the user. For each daily score response, we use keystroke information collected between 5AM on the previous day to 5AM on the current day. We then extracted the following features:

\textit{Text:} After removing stop words, we collected the top 1000 words (out of approximately 3.2 million) used across all users in our dataset and created a \textit{bag-of-words} feature that contains the daily number of occurrences of each word.

\textit{App usage:} We count the number of keystrokes entered per each application, creating a \textit{bag-of-apps} feature for each day. We discard applications that are used by less than 10\% of the participants so that our features are generalizable to more than just a single user in the dataset, leaving 137 totals apps (out of the original 640).

We observed that our models performed well when binarizing our feature vectors into boolean vectors, which signify whether a word or app was used on a given day (i.e. mapping values greater than 0 to 1). Our final feature vectors consist of a concatenation of a normalized and a binarized feature vector, resulting in 2000 and 274-dimensional vectors for Text and App features respectively.




\vspace{-1mm}
\section{Methods}
\vspace{-1mm}

In this paper we focus on studying approaches for learning privacy-preserving representations from mobile data for mood prediction. Our processed data comes in the form of triples $\{(x_{t,i}, x_{a,i}, y_i)\}_{i=1}^n$ with $x_t \in \mathbb{N}^{|V_t|}$ denoting the bag of text features with vocabulary size $|V_t| = 2000$ and $x_a \in \mathbb{N}^{|V_a|}$ denoting the bag of apps features with vocabulary size $|V_a| = 274$. $y$ denotes the label which takes on 3 categories. In parallel, we also have data representing the corresponding (one-hot) user identity $x_{\textrm{id}}$ which will be useful when learning privacy-preserving representations that do not encode information about $x_{\textrm{id}}$ (and evaluating privacy performance).

\textbf{Unimodal approaches:} We begin with a description of 2 unimodal approaches we considered.

1. Support Vector Machines (\textsc{SVMs}) project training examples to a chosen kernel space and finds the optimal hyperplane that maximally separates each class of instances. We directly apply a SVM classifier on input data $x_{\textrm{uni}} \in \{ x_t, x_a \}$ and train with standard supervised learning to predict daily mood labels $y$.

2. Multilayer Perceptrons (\textsc{MLPs}) have seen widespread success in supervised prediction tasks due to their ability in modeling complex nonlinear relationships. Because of the small size of our dataset, we choose a simple multilayer perceptron with two hidden layers. Similarly, we apply an \textsc{MLP} classifier on input data $x_{\textrm{uni}} \in \{ x_t, x_a \}$ to predict daily mood labels $y$.

\textbf{Multimodal models:} We extend both \textsc{SVM} and \textsc{MLP} classifiers using early fusion~\citep{baltruvsaitis2018multimodal} of text and app usage to model multimodal interactions. Specifically, we align the input through concatenating the bag-of-words and bag-of-apps features for each day before, resulting in an input vector $x_{\textrm{multi}} = x_t \oplus x_a$, before using a \textsc{SVM}/\textsc{MLP} classifier for prediction.

\begin{figure}%
    \centering
    \includegraphics[width=\textwidth]{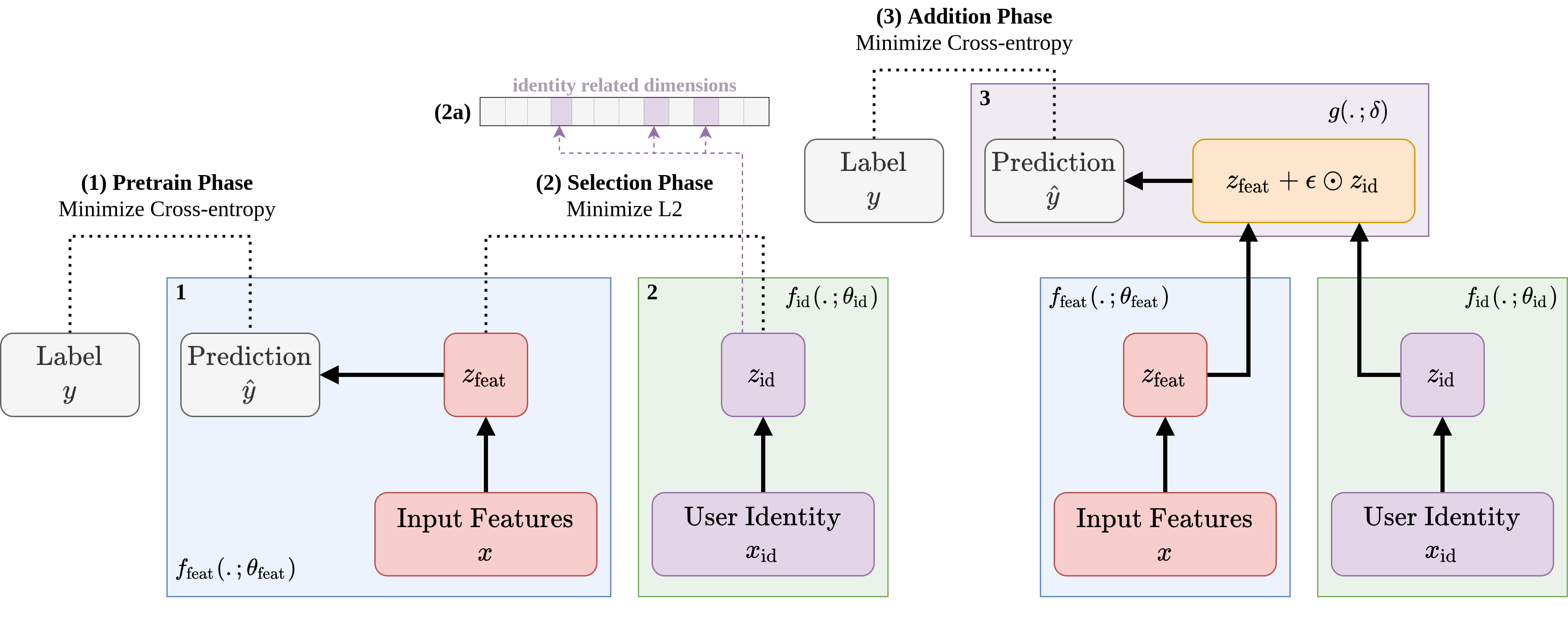}
    \caption{Diagram of the \names\ architecture. Boxes with numbers denote which parameters are being optimized in the corresponding step. For example in the addition phase (3), \names\ optimizes parameters $\delta$ in $g(.;\delta)$. 2a depicts $z_\textrm{id}$, which is a sparse vector of size $\text{dim} (z_\textrm{feat})$ whose nonzero values (colored purple) signify identity-confounding dimensions in $z_\textrm{feat}$.} 
    \label{fig:sal}%
\end{figure}

\textbf{A step toward preserving privacy:} While classifiers trained with supervised learning can learn useful representations for mood prediction, they carry the risk of \textit{memorizing} the identity of the user along with their sensitive mobile usage and baseline mood scores, and possibly \textit{revealing} these identities to adversarial third-parties. Therefore, we aim to achieve a balance between predictive performance of mood while protecting the privacy of personal identities.

We adapt the Selective-Additive Learning framework~\citep{wang2017select} with the goal of learning \textit{disentangled} representations separated into \textit{identity-dependent} and \textit{identity-dependent} factors using 2 phases:

\textit{Selection phase:} We are first presented with a set of (multimodal) features that are contain both identity-dependent and independent dimensions $z_\textrm{feat} = f_\textrm{feat}(x; \theta_\textrm{feat}^*)$ by pretraining a \textsc{MLP} classifier using supervised learning. $f_\textrm{feat}$ denotes the classifier with pretrained parameters $\theta_\textrm{feat}^*$ and $x$ denotes the input feature vector. Our goal is to now disentangle the identity-dependent and independent dimensions within $z_\textrm{feat}$, which can be done by training a separate classifier to predict $z_\textrm{feat}$ \textit{as much as possible} given only the user identity:
\begin{equation}
    \label{eq:SAL}
    \theta_\textrm{id}^* = \argmin_{\theta_\textrm{id}} \left( z_\textrm{feat} - f_\textrm{id} (x_\textrm{id}; \theta_\textrm{id}) \right)^2 + \lambda ||\theta_{\textrm{feat}}||_1,
\end{equation}
where $x_\textrm{id}$ denotes a one hot encoding of user identity, $f_\textrm{id}$ denotes the identity encoder with parameters $\theta_\textrm{id}$, and $\lambda$ denotes a hyperparameter that controls the weight of the L1 regularizer. $f_\textrm{id}$ projects the user identity encodings to the feature space learned by $f_\textrm{feat}$. By minimizing the objective in equation~\ref{eq:SAL} for each $(x, x_\textrm{id})$ pair, $f_\textrm{id}$ learns to encodes user identity into a sparse vector $z_\textrm{id} = f_\textrm{id} (x_\textrm{id}; \theta_\textrm{id}^*)$, whose nonzero values represent identity-dependent dimensions in $z_\textrm{feat}$.

\textit{Additive phase:} Given these 2 factors $z_\textrm{feat}$ and $z_\textrm{id}$ and to ensure that our prediction model does not capture identity-related information $z_\textrm{id}$, we add multiplicative Gaussian noise to the identity-related dimensions $z_\textrm{id}$ while repeatedly optimizing for mood prediction with a final \textsc{MLP} classification layer $g(z_\textrm{feat},  z_\textrm{id} ; \delta)$. This resulting model only retains identity-independent features for mood prediction:
\begin{equation}
    \hat{y} = g \left( z_\textrm{feat} + \epsilon \odot z_\textrm{id} \right)
\end{equation}
where $\epsilon \sim \mathcal{N}(0, \sigma^2)$ is repeatedly sampled across batches and training epochs. We call this approach \namel, or \names\ for short.

\textbf{Controlling the tradeoff between performance and privacy:} We found that directly applying noise to the identity-dependent dimensions led to improved privacy at the expense of prediction performance. As a result, to control the tradeoff between performance and privacy, we vary the parameter $\sigma$, which controls amount of noise added to the identity-dependent dimensions across batches and training epochs. $\sigma=0$ recovers a standard \textsc{MLP} classifier with good performance but reveals user identities, while large $\sigma$ effectively protects user identities but at the possible expense of mood prediction performance.

In practice, the optimal tradeoff between privacy and performance varies depending on the problem. For our purposes, we report in Table \ref{modalities_comparison} and \ref{privacy} results of models which achieve the highest performance-privacy ratio $R$ on the validation set, where $R = \frac{s_{\textsc{MLP}} - s_{\names}}{t_{\textsc{MLP}} - t_{\names}}$ is defined as the improvement in privacy per unit of performance lost. Here, $s$ is defined as the accuracy in the user prediction task and $t$ is defined as the F1 score on the mood prediction task.

\vspace{-1mm}
\section{Experiments}
\vspace{-1mm}

We perform experiments to test the utility of text and app features in predicting daily mood while keeping user identity private.

\textbf{Data splits:} Given that our data is longitudinal, we split our data into 10 partitions order chronologically by user. We do so in order to maintain independence between the train, validation, and test splits in the case where there is some form of time-level dependencies within our labels. For evaluating \textsc{SVM} and \textsc{MLP}, we run a nested k-fold cross validation (i.e. we perform 9-fold validation within 10-fold testing). For each test fold, we identify the optimal parameter set as the one that achieves highest mean validation score over the validation folds. We choose average F1 score as our validation metric because our classes are imbalanced. To evaluate \names, we use the best performing \textsc{MLP} model for each test fold as our base neural classifier.

\textbf{Results:} For our experiments we report the test accuracy and macro F1 score. Given the relatively low number of cross validation folds, use the Wilcoxon signed-rank test \cite{wilcoxon1992individual} at the 5$\%$ significance level for all significance testing. We make the following observations regarding our research questions.

\textbf{Observation 1: Text and app usage features are predictive of mood.} To evaluate how predictive bag-of-words and app usage features are, we run experiments using \textsc{SVM}, \textsc{MLP}, and \names. Because we have unbalanced classes, we choose a majority classifier (i.e. predict the most common class in the training set) for our baseline. Results are reported in Table \ref{modalities_comparison}. We observe that using our features, the models outperform the baseline with respect to accuracy and F1 score. Using the Wilcoxon test at the 5$\%$ significance level, we find that these improvements over the baseline in both F1 score and accuracy are statistically significant (p-value $<< 0.05$).

\definecolor{gg}{RGB}{15,125,15}
\definecolor{rr}{RGB}{190,45,45}

\begin{table*}[!tbp]
\fontsize{8.5}{11}\selectfont
\centering
\caption{Comparison of mood prediction performance across different modalities. Best results in \textbf{bold}. For both accuracy and F1 score, models trained on both text and apps features outperform models trained using individual modalities. $\star$ and $\diamond$ denote that the difference between multimodal and unimodal (Text and Apps respectively) is statistically significant (p-value $<< 0.05$).}
\setlength\tabcolsep{5.0pt}
\begin{tabular}{l || c c c c | c c c c }
\Xhline{3\arrayrulewidth}
& \multicolumn{4}{c|}{{F1 score}} & \multicolumn{4}{c}{{Accuracy}} \\
Modalities & \textsc{Baseline} & \textsc{SVM} & \textsc{MLP} & \names &  \textsc{Baseline} & \textsc{SVM} & \textsc{MLP} & \names \\
\Xhline{0.5\arrayrulewidth}
Text + Apps &  $19.07$ & $\textbf{62.08}^{\diamond}$ & $\textbf{58.38}^{\star\diamond}$ & $\textbf{52.90}$ & $40.18$ & $\textbf{66.59}^{\diamond}$ & $\textbf{62.93}^{\star\diamond}$ & $\textbf{56.76}$ \\
Text & $19.07$ & $61.15 $ & $56.27$ & $52.63$ & $40.18$ & $65.83$ & $60.61$ & $56.08$ \\
Apps & $19.07$ & $58.65$ & $52.29$ & $51.32$ & $40.18$ & $62.65$ & $55.26$ & $55.68$ \\
\Xhline{3\arrayrulewidth}
\end{tabular}
\label{modalities_comparison}
\end{table*}

\textbf{Observation 2: Fusing both text and app usage modalities improves performance.} In Table \ref{modalities_comparison}, we also compare the performance of our models on combined (Text + Apps) features to the performance on each individual feature set. We observe that for both metrics, combining the two features gives better performance over using either modality individually. 



\begin{figure}%
    \centering
    \subfloat[\centering \textsc{MLP}]{{\includegraphics[width=7cm]{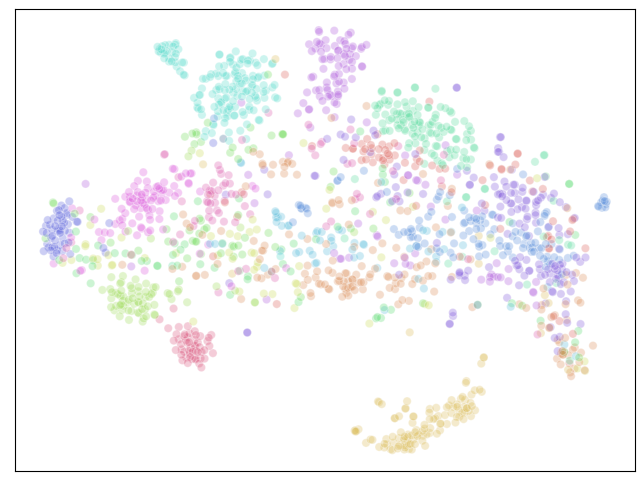} }}%
    \qquad
    \subfloat[\centering \names]{{\includegraphics[width=7cm]{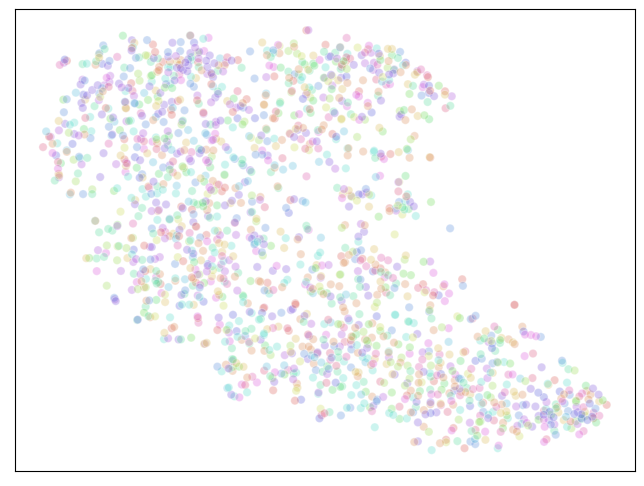} }}%
    \caption{Visualization of representations learned by (a) \textsc{MLP} and (b) \names, which have been reduced to two dimensions via t-SNE and colored by participant identity. Representations learned by \names\ are no longer separable by user which better preserve privacy.} 
    \label{fig:privacy_representations}%
\end{figure}

\textbf{Observation 3: Learning privacy-preserving features} While text and app history features are predictive of mood, we observe that such features are heavily correlated with user identity through both visual and empirical evaluations:

\textit{Visual comparison.} We use t-SNE to reduce the learned features to 2 dimensions and after color-coding the points by participant identity, we see distinct clusters in Figure \ref{fig:privacy_representations}(a). After training \names, we again visualize the representations learned in Figure \ref{fig:privacy_representations}(b) and we find that they are no longer visually separable by user, indicating that \names\ indeed learns user-independent feature representations.

\textit{Quantitative evaluation.} To empirically evaluate how well our models preserve privacy, we extracted each trained model's penultimate layers and trained a logistic regression model to predict user identity from extracted feature representations. The more a model preserves privacy, the harder it should be to predict user identity. We observed that we can predict user identity based on the raw features with near perfect accuracy ($95\%$ using Text + Apps) which indicate that the labels are highly correlated with user identities. Next, we evaluated how each trained model's feature representations encode information about user identities. While a logistic regression model can achieve up to $80\%$ accuracy in predicting users from embeddings learned by the \textsc{MLP}, it achieves significantly lower accuracy when trained on \names\ embeddings - only $37\%$.

Note that we do not include results for SVM, which finds a linear separator in a specified kernel space rather than learning a representation for each sample. Moreover, explicitly projecting our features is computationally infeasible due the dimensionality of our chosen kernel spaces. 

\textit{Understanding the tradeoff between performance and privacy.} \names\ provides a tunable parameter $\sigma$ to control the privacy of user identities while remaining predictive of daily mood. In Table~\ref{privacy} (right) we plot the tradeoff between performance (mood prediction F1 score, higher is better) and privacy (identity prediction accuracy, lower is better). We find that using a multimodal model on both text and app features pushes the Pareto front to obtain better performance and privacy at the same time. Interestingly, we find that the tradeoff curve on a model trained only using app features does not exhibit a Pareto tradeoff curve as expected. We attribute this to randomness in predicting both mood and identities. Furthermore, Wang et al.~\cite{wang2017select} found that adding noise to identity dimensions can sometimes improve generalization by reducing reliance on identity-confounding dimensions, which could also explain occasional increased performance at larger $\sigma$ values.

\definecolor{gg}{RGB}{15,125,15}
\definecolor{rr}{RGB}{190,45,45}

\begin{table*}[t!]
\begin{minipage}{.45\linewidth}
\fontsize{9}{11}\selectfont
\caption{\textbf{Left:} We evaluate privacy in predicting user identity from representations learned by \textsc{MLP} and \names\ (\textbf{lower} accuracy is better), and find that \names\ effectively obfuscates user identity. \textbf{Right:} Tradeoff between performance (mood prediction F1 score, \textbf{higher} is better) and privacy (identity prediction accuracy, \textbf{lower} is better) \names\ provides a tunable parameter $\sigma$ which allows us to plot a range of (performance, privacy) points. Furthermore, using a multimodal model on text and app features pushes the Pareto front to obtain better performance and privacy at the same time.}
\setlength\tabcolsep{6.0pt}
\begin{tabular}{l || c c c }
\Xhline{3\arrayrulewidth}
& Text + Apps & Text & Apps \\
\Xhline{0.5\arrayrulewidth}
Raw features & $95.70$ & $92.65$ & $91.82$ \\ 
\textsc{MLP} & $79.04$ & $76.41$ & $85.94$ \\ 
\names & $\textbf{36.65}$ & $\textbf{38.38}$ & $\textbf{36.72}$ \\
\Xhline{3\arrayrulewidth}
\end{tabular}
\label{privacy}
\end{minipage}
\hspace{4mm}
\vspace{-4mm}
\begin{minipage}{.55\linewidth}
\centering
\includegraphics[width=\textwidth]{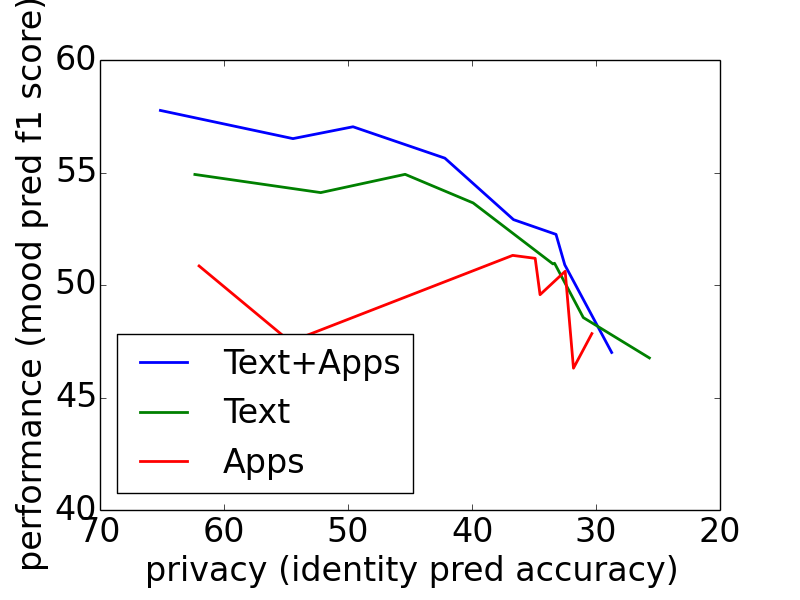}
\end{minipage}

\vspace{2mm}
\end{table*}

\textbf{Additional observations:} Since this is a new dataset, we explored several more machine learning methods throughout the research process. In this section we describe some of the approaches that yielded initial negative results despite them working for standard datasets in machine learning and natural language processing:
\begin{enumerate}
    \item \textbf{Pretrained word and sentence representations:} We initially tried finetuning language model (BERT) and using word embedding (GLoVe) methods to extract text features, classical approaches which work extremely well for NLP problems. Surprisingly, we were not able to outperform classifiers based on bag-of-words features. We hypothesize that it is difficult to use these methods to capture such long sequences of data spread out over a single day. One avenue we wish to explore is to treat this feature extract as a ranking problem. This proposal is based on the assumption that in the course of a day, only a few key events/periods of time are predictive of mood, forcing us to design a method that identifies them.
    \item \textbf{Keystroke features:} We also experimented with keystroke/typing pattern features by extracting several summary statistics as features (e.g. average number of words typed per minute), but these did not improve the model when combined with text and app features. Both the encoder for keystroke data as well as the better multimodal fusion with text and app features could help in this regard. For example, fine-grained fusion that maps each keystroke to the application in which the text was entered in~\citep{liang2018multimodal}.
    \item \textbf{User specific models:} We also explored the setting of training a separate model per user but we found that there was too little data per user to train a good model. As part of future work, we believe that if \names\ can learn a user independent neural classifier, this approach can be used pretrained representations for either finetuning or few shot learning on each specific user. Previous work in federated learning~\citep{liang2020think,smith2017federated} also offer ways of learning a user-specific model that leverages other users' data during training, which could help to alleviate the lack of data per user.
    \item \textbf{User-independent data splits:} We have shown that text and app features are highly dependent on participant identities. Consequently, models trained on these features would perform poorly when evaluated on a user not found in the training set. We would like to evaluate if better learning of user independent features can improve generalization to new users (e.g. split the data such that the first 10 users are used for training, next 3 for validation, and final 4 for testing). 
\end{enumerate}

\vspace{-1mm}
\section{Conclusion}
\vspace{-1mm}

In this paper, we used multimodal machine learning to learn markers of daily mood from mobile data. Our method is also able to obfuscate user identity as a step towards more privacy-preserving learning, a direction crucial towards real-world learning from sensitive mobile data. Several important future directions involve: 1) investigating the relationship between user-dependent and independent features to characterize the tradeoffs between performance and privacy, 2) improving multimodal representations through more fine-grained temporal fusion~\citep{liang2018multimodal}, 3) incorporating more unique features provided in the dataset including keystrokes and geolocation data, and 4) federated training from decentralized data~\citep{DBLP:journals/corr/McMahanMRA16} to further improve privacy~\citep{geyer2017differentially} and fairness~\citep{liang2020fair} of sensitive mobile data.

\vspace{-1mm}
\section*{Acknowledgements}
\vspace{-1mm}

This material was based upon work partially supported by the National Science Foundation (Award \#1722822) and National Institutes of Health (Award \#U01MH116923). MM was supported by the Swiss National Science Foundation (\#P2GEP2\_184518). Any opinions, findings, and conclusions or recommendations expressed in this material are those of the author(s) and do not necessarily reflect the views of National Science Foundation or National Institutes of Health, and no official endorsement should be inferred.

\bibliography{refs}
\bibliographystyle{plain}


\section*{Appendix}

\section{Implementation Details}
\label{implementation_details}

\textbf{Hyperparameters:} We show all hyperparameters used in Tables~\ref{params}.

\begin{table*}[!tbp]
\fontsize{8.5}{11}\selectfont
\centering
\caption{Model parameter configurations. \\
*Integer kernel values denote the degree of a polynomial kernel.}
\setlength\tabcolsep{6.0pt}
\begin{tabular}{l | c | c }
\Xhline{3\arrayrulewidth}
Model & Parameter & Value \\
\Xhline{0.5\arrayrulewidth}
\multirow{2}{*}{SVM}
& C & 0.1, 0.5, 1, 2, 3, 5, 10 \\
& Kernel* & RBF, 2, 3, 5, 10 \\
\Xhline{0.5\arrayrulewidth}
\multirow{6}{*}{MLP}
& hidden dim 1 & 1024, 512, 128 \\
& hidden dim 2 & 128, 64 \\
& dropout rate & 0, 0.2, 0.5 \\
& learning rate & 0.001 \\
& batch size & 100 \\
& epochs & 200 \\
\Xhline{0.5\arrayrulewidth}
\multirow{2}{*}{\names}
& $\lambda$ & 0.1, 1, 2, 3, 5, 10 \\
& $\sigma$ & 1, 5, 10, 25, 50, 100 \\
\Xhline{3\arrayrulewidth}
\end{tabular}
\label{params}
\end{table*}

\end{document}